\documentclass[conference]{IEEEtran}
\usepackage[ sorting=none]{biblatex}
\usepackage{xcolor}
\usepackage{graphicx}%
\usepackage{multirow}
\usepackage{tabularx}
\usepackage{afterpage} % For delaying figure placement

\ifCLASSINFOpdf
\else
\fi

\begin{document}
%
% paper title
% Titles are generally capitalized except for words such as a, an, and, as,
% at, but, by, for, in, nor, of, on, or, the, to and up, which are usually
% not capitalized unless they are the first or last word of the title.
% Linebreaks \\ can be used within to get better formatting as desired.
% Do not put math or special symbols in the title.
\title{Synthesizing CTA Image Data for Type-B Aortic Dissection using Stable Diffusion Models}

% author names and affiliations
% use a multiple column layout for up to three different
% affiliations
% \author{\IEEEauthorblockN{Ayman Abaid}
% \IEEEauthorblockA{School of Electrical and\\Computer Engineering\\
% Georgia Institute of Technology\\
% Atlanta, Georgia 30332--0250\\
% Email: http://www.michaelshell.org/contact.html}
% \and
% \IEEEauthorblockN{Simpson}
% \IEEEauthorblockA{Twentieth Century Fox\\
% Springfield, USA\\
% Email: homer@thesimpsons.com}
% \and
% \IEEEauthorblockN{James Kirk\\ and Montgomery Scott}
% \IEEEauthorblockA{Starfleet Academy\\
% San Francisco, California 96678--2391\\
% Telephone: (800) 555--1212\\
% Fax: (888) 555--1212}}

% conference papers do not typically use \thanks and this command
% is locked out in conference mode. If really needed, such as for
% the acknowledgment of grants, issue a \IEEEoverridecommandlockouts
% after \documentclass

% for over three affiliations, or if they all won't fit within the width
% of the page, use this alternative format:
% 
\author{\IEEEauthorblockN{Ayman Abaid\IEEEauthorrefmark{1},
 Muhammad Ali Farooq\IEEEauthorrefmark{2},
Niamh Hynes\IEEEauthorrefmark{4},
Peter Corcoran\IEEEauthorrefmark{2}, and
Ihsan Ullah\IEEEauthorrefmark{1}\IEEEauthorrefmark{2}\IEEEauthorrefmark{3}} 

Email:\{a.abaid1,muhammadali.farooq,niamh.hynes,peter.corcoran,ihsan.ullah\}@universityofgalway.ie 
\IEEEauthorblockA{\IEEEauthorrefmark{1}School of Computer Science, University of Galway, Ireland\\
\IEEEauthorrefmark{2} School of Engineering, College of Science and Engineering,
University of Galway, Ireland\\
\IEEEauthorrefmark{3}Insight SFI Research Center for Data Analytics, University of Galway, Ireland}
\IEEEauthorrefmark{4}University Hospital Galway, Newcastle Road, University of Galway, Ireland}
% \IEEEauthorblockA{\IEEEauthorrefmark{2}Twentieth Century Fox, Springfield, USA\\
% Email: homer@thesimpsons.com}
% \IEEEauthorblockA{\IEEEauthorrefmark{3}Starfleet Academy, San Francisco, California 96678-2391\\
% Telephone: (800) 555--1212, Fax: (888) 555--1212}
% \IEEEauthorblockA{\IEEEauthorrefmark{4}Tyrell Inc., 123 Replicant Street, Los Angeles, California 90210--4321}
% }

% use for special paper notices
% use for special paper notices
%\IEEEspecialpapernotice{(Invited Paper)}

% make the title area
\maketitle

% As a general rule, do not put math, special symbols or citations
% in the abstract
\begin{abstract}
Stable Diffusion (SD) has gained a lot of attention in recent years in the field of Generative AI thus helping in  synthesizing medical imaging data with distinct features. The  aim is to contribute to the ongoing effort focused on overcoming the limitations of data scarcity and improving the capabilities of ML algorithms for cardiovascular image processing. Therefore, in this study, the possibility of generating synthetic cardiac CTA images was explored by fine-tuning stable diffusion models based on user defined text prompts, using only limited number of CTA  images as input. A comprehensive evaluation of the synthetic data was conducted by incorporating both quantitative analysis and qualitative assessment, where a clinician assessed the quality of the generated data. It has been shown that Cardiac CTA images can be successfully generated using using Text to Image (T2I) stable diffusion model. The results demonstrate that the tuned T2I CTA diffusion model was able to generate images with features that are typically unique to acute type B aortic dissection (TBAD) medical conditions.
% Stable Diffusion (SD) has gained a lot of attention in recent years in the field of Generative AI thus helping in  synthesizing medical imaging data with distinct features. The  aim is to contribute to the ongoing effort focused on overcoming the limitations of data scarcity and improving the capabilities of ML algorithms for cardiovascular image processing. Therefore, in this study, the possibility of generating synthetic cardiac CTA images was explored by fine-tuning stable diffusion models based on user defined text prompts, using only limited number of CTA  images as input. A comprehensive evaluation of the synthetic data was conducted by incorporating both quantitative analysis and qualitative assessment, where a clinician assessed the quality of the generated data. To the best of our knowledge, this is the first attempt to generate cardiac CTA images using Text to Image (T21) stable diffusion model. The results demonstrate that the tuned T2I CTA diffusion model was able to generate images with features that are typically unique to acute type B aortic dissection (TBAD) medical conditions.

\end{abstract}

% no keywords

% For peer review papers, you can put extra information on the cover
% page as needed:
% \ifCLASSOPTIONpeerreview
% \begin{center} \bfseries EDICS Category: 3-BBND \end{center}
% \fi
%
% For peerreview papers, this IEEEtran command inserts a page break and
% creates the second title. It will be ignored for other modes.
\IEEEpeerreviewmaketitle

\section{Introduction}
% no \IEEEPARstart
Aortic dissection (AD) occurs when there is a tear in the inner layer of the aorta, the largest artery of the heart. The Stanford classification system divides AD into two types depending upon the localization of the dissection. Type B aortic dissection (TBAD) specifically occurs when dissection takes place  in the descending aorta. This tear results in the separation of the inner and middle layers of the aorta, creating a second pathway for blood flow known as a false lumen (FL), which can also obstruct blood flow in the true aortic channel, i.e., the true lumen (TL) \cite{juang2008aortic}. Computed Tomography Angiography (CTA) is the most widely used modality for diagnosing TBAD due to its widespread availability, precision, and ability to provide quick results \cite{braverman2011aortic}.
% The diagnosis of aortic dissection often relies on CT scanning, MRI, and transesophageal echocardiography, all of which are reasonably accurate modalities. Among these modalities, 

The diagnostic and prognostic process of TBAD involves the segmentation of FL, TL, and the  false lumen thrombosis (FLT). Manual segmentation of TL, FL, and FLT is challenging and time-intensive. Previous studies e.g. \cite{yao2021imagetbad,wobben2021deep} have employed deep learning (DL) based models to automate the segmentation process, reporting promising results on testing set. Specifically, they achieved mean Dice similarity coefficient (DSC) score ranging from 0.85-0.93 for TL, 0.78-0.91 for FL and 0.50-0.52  for FLT respectively. Given the low incidence of TBAD, a significant constraint arises from the limited size of the dataset employed for training and testing. This limitation becomes particularly noticeable when dealing with instances of uncommon morphologies, such as FLT, and it hampers the optimal performance of the DNN model \cite{hahn2020ct}. Since the availability of large-scale labeled data is a contributing factor to the success of DL models, there is a significant need to gather data with varying morphologic findings and imaging parameters \cite{sun2017revisiting}.
% \textcolor{blue}{Do I need to mention which image modalities have been synthetically generated by AI(GAN or SD) till today?
% no just focus on Cardiovascular modality and see if research community has worked on generating data via GANS/ SD etc ..}
% anatomical features linked to Type B Aortic Dissection.

Generative AI plays a significant and evolving role in medical imaging across various applications. Generative models can create synthetic labeled medical images, helping to augment limited datasets. This is particularly valuable when dealing with rare conditions or when labeled datasets are scarce. Further synthetic datasets introduce diversity in the existing dataset, aiding in the training of deep neural network (DNN) models that can handle variations in patient demographics and conditions. In the domain of cardiac CTA imaging, the integration of AI-based generative algorithms specifically probabilistic diffusion models \cite{croitoru2023diffusion} marks a significant advancement. Such type of State-of-the-Art (SoTA) models can render synthetic images with improved stability and authenticity, mimicking the intricate details of the cardiovascular system. The stability introduced by diffusion models ensures the representation of complex anatomical structures and physiological processes.

Keeping this in view, we have incorporated Text-to-Image (T2I) diffusion models \cite{wu2023uncovering} as an innovative approach in the realm of medical imaging, particularly for conveying textual clinical information into visually realistic images. These models leverage diffusion processes to simulate the stepwise transformation of a latent representation into a coherent CTA representation which is guided via textual descriptions. These synthesized images can play a crucial role in tasks such as image segmentation, disease detection, identifying cardiac abnormalities, and optimizing treatment strategies

\begin{figure*}[t]
  \centering
  \includegraphics[width=0.9\textwidth, height=8cm] {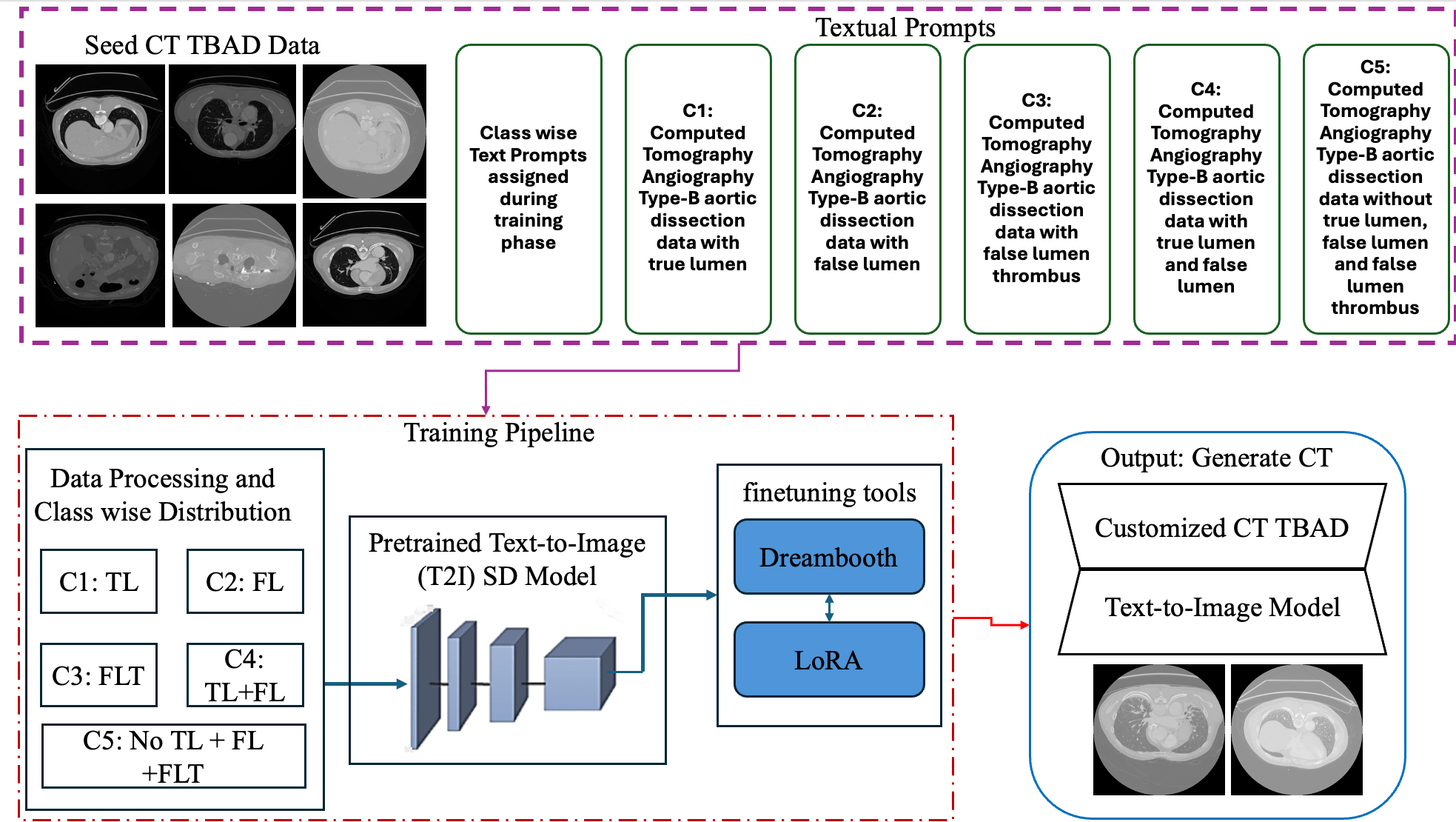}
  \vspace{-0.5em}
  \caption{Methodology for training SD models and generating CTA images via diverse image sampling}
  \label{fig:stable_diffusion_block_diagram}
\end{figure*}
The key contributions of this study are as follows:
\begin{enumerate}
  \item  Clinical data understanding, acquisition, and pre-processing TBAD images.
  \item Optimal tuning of stable diffusion model for synthesizing TBAD images by using only limited number of seed data and incorporating low rank adaption (LoRA) model.
  \item Rendering TBAD data with advanced data augmentation based on user specific text prompts, thereby showing explicit details of TL, FL, FLT and the combination of TL and FL.
  \item Extensive quantitative \& qualitative validation of CTA rendered data, incorporating the opinions of medical experts.
\end{enumerate}

\begin{figure*}[t]
  \centering
  \includegraphics[width=0.9\textwidth , height=8cm] {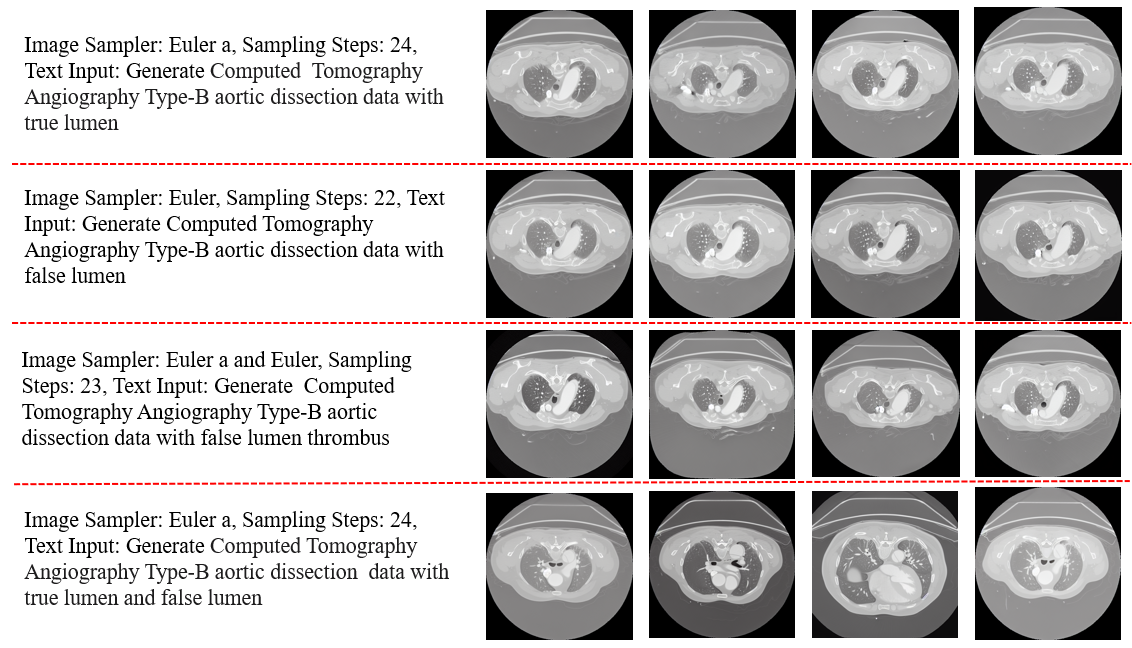}
  \vspace{-1.2em}
  \caption{Rendered CT data and advanced data augmentation results with distinct TL, FL, FLT and TL + FL features }
  \label{data_augmentation}
\end{figure*}
\section{Proposed Methodology}
For this study, we utilized the ImageTBAD dataset \cite{yao2021imagetbad}, comprising 100 3D CTA images annotated for FL, TL, and FLT. In the pre-processing step, we converted the 3D images into 2D images by considering each slice as an individual image. We partitioned the dataset into training, testing, and validation sets. Furthermore, we divided data into five main classes i.e. class 1: images with TL, class 2: images with FL, class 3: images with FLT,
class 4: images with TL \& FL, and class 5: data without any TL, FL and FLT information. In the training set, Class 1 comprises 501 samples, Class 2 is represented by 387 samples, Class 3 exhibits 121 samples, Class 4 has 13,544 samples, while Class 5 contributes a set of 3,582 samples.
%  our dataset comprises five distinct classes, each with varying sample size

% Specifically, 23,844 images were used for training, 5,300 for validation, and 5,236 for testing. 

\label{sec2}
Further to increase the existing TBAD data and indulge extensive data diversity with specific information of FL/TL/FLT we have proposed a customized text-to-image (T2I) model for rendering high quality CTA data by using only limited number of seed data representation. In the context of medical imaging, this is especially crucial where acquiring extensive labelled datasets for training is challenging due to factors such as privacy concerns or the rarity of certain medical conditions. Conventional image generation models encounter challenges when seed data is limited, lacking an effective mechanism to seamlessly integrate new data classes into the training process. Consequently, issues like model overfitting arise, where the model replicates the exact input images, and language drift occurs, leading to the generation of undesired outputs due to semantic confusion. Fig. \ref{fig:stable_diffusion_block_diagram} shows the complete block diagram representation of adapted methodology for training SD models and further rendering CTA images by using various image sampling methods. 

The first stage includes acquiring the seed data and including it in
the training pipeline for tuning pretrained stable diffusion models. 
For synthetic image generation  only  data from training set was used, ensuring no information leakage across sets.
The overall data was pre-processed by bringing it to 512X512 image dimensions. In the second stage, keeping in view the class distinctions, we assigned specialized text prompts to the data of each class and further provided negative prompts to specific class of subsequent classes. The inclusion of negative prompts during the training phase is crucial for enhancing the robustness and reliability of diffusion models. It helps in identifying and mitigating potential biases, improving the model's ability to handle diverse user inputs, and ensuring that the system renders data appropriately as per the user expectations. %Figure \ref{data_augmentation} shows the data samples of each of these classes.

For fine tuning pretrained diffusion models, we integrated dreambooth training tool \cite{ruiz2023dreambooth} developed by Google Brain team. DreamBooth is a training technique which addresses a research gap by overcoming the limitations of existing text-image diffusion models. Although prior models like GANs have exhibited success in introducing specific class during training, they are often confined to specific domains and demand a substantial number of training samples. The goal of the DreamBooth technique is to establish a more efficient  fine-tuning method by using only limited number of images that retains semantic class knowledge, ultimately facilitating effective subject-driven generation. 
The fine-tuning approach implemented in the DreamBooth technique utilizes the pre-trained diffusion model, Imagen \cite{saharia2022photorealistic}, along with a unique token identifier to introduce a new subject. Through the fine-tuning process involving the rare token identifier and particular loss function, the authors have demonstrated effective training strategy to produce high-quality images of the novel class. This method effectively retains the prior knowledge embedded in the pre-trained model, enabling efficient learning with limited seed data. During the training phase, we experimented with various combinations of  network hyper-parameters to find the optimal configuration for our specific task and dataset. The choice of hyperparameters can have a significant impact on the performance and model convergence. In our case, the total epochs were 100 with batch size of 2, while the base learning rate was set to 1e-4. We opted for the 8-bit AdamW optimizer and used the LoRA method to cut the computational cost. Additionally, we selected the Discrete Denoising Scheduler (DDS) as the noise scheduler, and the tuned model checkpoint was stored in half-precision fp-16 to  significantly reduced the VRAM footprint and increase the model inference speed. 

We validated the effectiveness of the proposed method by statistically analysing the quality and diversity of synthetic images for each class. For this purpose we used following metrics: Fréchet Inception Distance (FID) \cite{heusel2017gans} and Multiscale Structural Similarity Index Measure (MS-SSIM) \cite{wang2003multiscale}. Following the approach in  previous studies \cite{dhariwal2021diffusion,muller2022diffusion}, we utilized Inception-v3 to extract features. By comparing the features derived from the deepest layer of the Inception-v3 model, we computed the distance between the distributions of real images from the test set and synthetic images.
MS-SSIM  \cite{wang2003multiscale} is an extension of SSIM \cite{wang2004image} that measures the structural similarity between two images by taking luminance, contrast, and structural information into account. To calculate MS-SSIM, we randomly selected pairs of images from the testing set and synthetic set for each class, then averaged the scores across all pairs. Lastly, to ensure the utility of synthetic images, we trained SoTA model i.e. UNet \cite{ronneberger2015u} on real data, which achieved high accuracy (DSC score of 0.83 for TL and  0.84 for FL) on the test dataset. Subsequently, we evaluate the quality of synthetic CTA images by subjecting them to analysis using our model, examining the presence of features that can be detected by the model.

% While there exists some skepticism within a segment of the research community \cite{jayasumana2023rethinking,chong2020effectively} regarding the use of Fréchet Inception Distance, it remains a widely adopted standard metric for quality comparisons. 
\section{Experimental Outcomes }
The experimental environment was setup on server grade machine equipped with Intel Xeon processor, and A6000 GPU card with 48GB of dedicated video memory.

\subsection{Data Rendering Results using Tuned TBAD T2I Model}
The overall training process was completed in four days. Incorporation of  LoRA \cite{hu2021lora} during the fine-tuning phase helps to train the systems on different concepts which includes 5 different classes of TBAD data as mentioned in Section~\ref{sec2}. Furthermore, LoRA enhances training efficiency and significantly reduces the hardware cost by using just 4.6 GB of video memory, especially with adaptive optimizers, thus achieving up to three-fold improvement.
% \section{Results \& Analysis}

The tuned model was stored in both safetensor and pytorch checkpoint format. Fig.\ref{data_augmentation} shows the generated CTA image samples by passing the model through the inference pipeline. These image samples were generated by using two different type of image samplers which includes euler and euler A with sampling steps ranging from 20-26. Selection of image samplers contribute to the production of diverse and realistic images by exploring different regions of the latent space. By sampling from various points, the tuned model can generate a wide range of outputs, capturing the diversity inherent in the training/seed data. In the next phase, we rendered CT data with class-wise distribution by using unique text prompts as shown in Fig.\ref{fig:stable_diffusion_block_diagram}. This approach helped in augmenting data with specific CT features, facilitating the distinction of TL, FL, and FLT in newly generated data. Further Fig.\ref{data_augmentation} elaborates the advanced data augmentation results containing TL, FL, FLT, and TL with FL features.

% \begin{figure}[thpb]
%     \centering
% \includegraphics[width=\columnwidth, height=6cm/Overall Rendered Data Figure-Ali .png}
% % \includegraphics[scale=0.35]
%     \caption{Sample CT Scans generated via tuned CT TBAD T2I stable diffusion model.}
%     \label{fig_4}
% \end{figure}

\begin{figure}
    \centering
\includegraphics[width=\columnwidth, height=6cm]{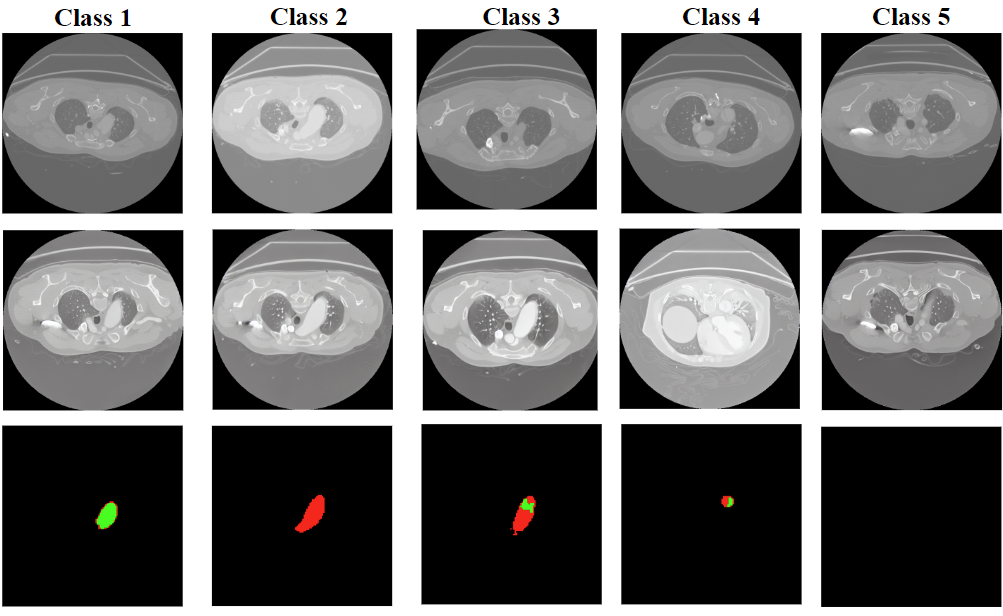}
\vspace{-2.5em}
    \caption{Real images (first row), synthetic images (second row) and segmentation results predicted by UNet on corresponding synthetic image (third row). In the segmented images, the color green corresponds to TL and red represents FL.}
    \label{fig:synthetic_real_unet}
\end{figure}

% \begin{figure*}[t]
%   \centering
%   % \includegraphics[width=\textwidth] {figures/temp.png}
%   \caption{Advanced data augmentation results with distinct TL, FL, FLT and TL + FL features }
%   \label{fig_5}
% \end{figure*}

% \afterpage{%
\begin{figure*}[t]
  \centering
  \includegraphics[width=\textwidth] {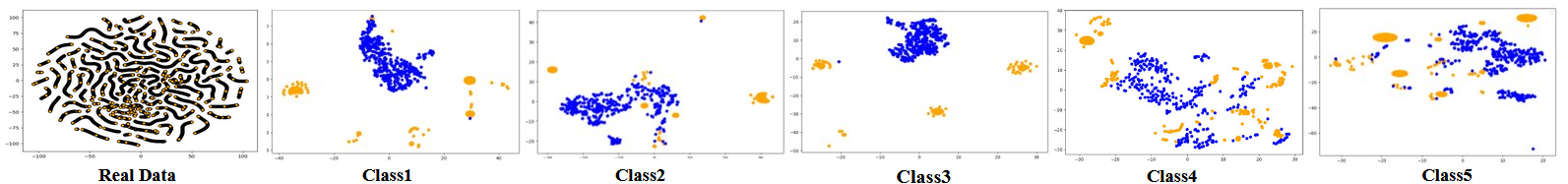}
    \vspace{-2.2em}
  \caption{t-SNE embedding displaying generated samples (shown in blue) from each class along with their nearest real CTAs (shown in orange).}
  \label{t_SNE}
\end{figure*}

% }
\subsection{Synthetic Data Validation}
Upon manual inspection by an untrained eye, the synthetic images exhibit a notable degree of realism and demonstrate superior fidelity in comparison to the original CTAs. However, it is noteworthy that the model produced few noisy images for class 1 and 2, yet remaining images of these classes retained a realistic quality in all other instances.

To measure structural similarity of real and synthetic images, MS-SSIM is computed for both real and synthetic images for each class. Higher MS-SSIM scores implies that two images are more similar,  while lower MS-SSIM scores signify a greater dissimilarity. Results reveal that the output from diffusion model consistently attained MS-SSIM scores closer to those of the original data across nearly all classes. Additionally, Fig. \ref{t_SNE} provides support for this claim by displaying the t-SNE embedding of synthetic CTA (shown in blue) alongside their nearest real counterparts (shown in orange) based on MS-SSIM for each class. The first chart showcases only real data, and owing to the three dimensional nature of data and the presence of multiple 2D images per patient, there is a significant overlap among the images corresponding to each patient. Despite relatively high FID score as compared to AI-generated natural images, it is important to note that fidelity measurement does not always correlate with utility  \cite{xing2023beauty} and further investigation is needed to evaluate their utility in performing downstream tasks. Results of the FID and MS-SSIM scores on our test set are reflected in Table \ref{tab:evaluation_metrics}. 
When the synthetic data was subjected to the UNet model trained on real data, successful detection of FL was achieved for class 2 and 4. However, for class 3, as illustrated in Fig. \ref{fig:synthetic_real_unet}, where the synthetic image was intended to contain FLT, the model failed to identify FLT. Given the inherently low segmentation accuracy of the model for FLT, any remarks on the quality of the synthetic image in this context is precluded.
Lastly, it is important to note that the metrics employed in this study for assessing image quality were not specifically designed for medical images. The development of metrics that correlates with human judgment remains an ongoing area of research \cite{borji2022pros}.

\section{Conclusion and Future Work}

This study highlights the potential of using customized T2I stable diffusion model to generate high quality synthetic CTA images, thus aiming to enhance the diagnosis of TBAD and acting as a building block for prognosis in the future. The rendered data was further used to augment the existing data, bringing diversity to the dataset and addressing prevalent issues of data scarcity. This includes addressing occurrences of rare morphologies like FLT and addressing privacy concerns within the medical domain. We showcase, by incorporating various quantitative metrics and qualitative evaluation, that the rendered CTA images exhibit fidelity equivalent to the real training data and represent their respective medical conditions to some extent. As the possible future directions, we aim to assess the utility of synthetic images in detail by evaluating their performance in downstream segmentation tasks. Furthermore, based on the clinician review, our emphasis will be on utilising clinical data alongside seed CTA videos to generate synthetic 3D CTA data by exploring advanced video diffusion models and assessing their quality and utility in the prognosis of TBAD.

%Furthermore, our focus will be on understanding how to appropriately assess the quality of synthetic medical images with the help of medical experts.

% \color{yellow}{Furthermore, our emphasis will be on producing 3D synthetic CTA images and assessing their utility in the prognosis of  TBAD, in alignment with the current clinical practice}
%\colorbox{yellow}{Furthermore, our emphasis will be on producing 3D CTA synthesis data by exploring advanced stable video diffusion models and assessing their utility in the prognosis of TBAD, and assessing the quality of synthetic medical images with the help of medical experts.}
%\colorbox{yellow}{ data by exploring advanced stable video diffusion models and assessing their utility in the prognosis of }
%\colorbox{yellow}{ TBAD, and assessing the quality of synthetic medical images with the help of medical experts.}
%aligning with requirements of the current clinical practice.}

 % Similarly, we have not yet attempted to synthesize labels 

% conference papers do not normally have an appendix

% use section* for acknowledgment
\section*{Acknowledgment}
The first author would like to thank the research funding from the College of Science and Engineering. In addition, this work is supported by ADAPT - Centre for Digital Content 
Technology, Enterprise Ireland, and with the financial support of Science Foundation Ireland under Grant Agreement No SFI/12/RC/2289\_P2.

\begin{table}
\centering
\caption{Evaluation Metrics: C1 corresponds to Class 1, C2 to Class 2, C3 to Class 3, C4 to Class 4, and C5 to Class 5}
\label{tab:evaluation_metrics}
\begin{tabular}{|c|ccccc|c|}
\hline
\multirow{2}{*}{} &
  \multicolumn{5}{c|}{\textbf{MS-SSIM}} &
  \multirow{2}{*}{\textbf{\begin{tabular}[c]{@{}c@{}}FID\\ (n=250)\end{tabular}}} \\ \cline{2-6}
 &
  \multicolumn{1}{c|}{\textbf{C1}} &
  \multicolumn{1}{c|}{\textbf{C2}} &
  \multicolumn{1}{c|}{\textbf{C3}} &
  \multicolumn{1}{c|}{\textbf{C4}} &
  \textbf{C5} &
   \\ \hline
\textbf{Real} &
  \multicolumn{1}{c|}{0.30} &
  \multicolumn{1}{c|}{0.32} &
  \multicolumn{1}{c|}{0.33} &
  \multicolumn{1}{c|}{0.31} &
  0.31 &
  \multirow{2}{*}{77.82} \\ \cline{1-6}
\textbf{Synthetic} &
  \multicolumn{1}{c|}{0.39} &
  \multicolumn{1}{c|}{0.37} &
  \multicolumn{1}{c|}{0.37} &
  \multicolumn{1}{c|}{0.38} &
  0.35 &
   \\ \hline
\end{tabular}
\end{table}

% trigger a \newpage just before the given reference
% number - used to balance the columns on the last page
% adjust value as needed - may need to be readjusted if
% the document is modified later
%\IEEEtriggeratref{8}
% The "triggered" command can be changed if desired:
%\IEEEtriggercmd{\enlargethispage{-5in}}

% references section
% \addbibresource{bibtex/IEEEexample.bib}

% can use a bibliography generated by BibTeX as a .bbl file
% BibTeX documentation can be easily obtained at:
% http://mirror.ctan.org/biblio/bibtex/contrib/doc/
% The IEEEtran BibTeX style support page is at:
% http://www.michaelshell.org/tex/ieeetran/bibtex/
% \bibliographystyle{IEEEtran}
% argument is your BibTeX string definitions and bibliography database(s)
% \bibliography{bibtex/IEEEexample}
%
% <OR> manually copy in the resultant .bbl file
% set second argument of \begin to the number of references
% (used to reserve space for the reference number labels box)
% \end{thebibliography}

\printbibliography

% that's all folks
\end{document}